# Improving Customer Service with Automatic Topic Detection in User Emails

Bojana Bašaragin[1][0000-0002-7679-1676] and Darija Medvecki[1][0000-0002-4180-0050] and Gorana Gojić[1][0000-0002-6100-5871] and Milena Oparnica[1][0000-0002-5986-3068] and Dragiša Mišković[1][0000-0002-0455-9552]

[1] The Institute for Artificial Intelligence Research and Development of Serbia, Fruškogorska 1, 21000 Novi Sad, Serbia
{bojana.basaragin,darija.medvecki, gorana.gojic,milena.oparnica,dragisa.miskovic}@ivi.ac.rs

**Abstract.** This study introduces a novel natural language processing pipeline that enhances customer service efficiency at Telekom Srbija, a leading Serbian telecommunications company, through automated email topic detection and labeling. Central to the pipeline is BERTopic, a modular framework that allows unsupervised topic modeling. After a series of preprocessing and postprocessing steps, we assign one of 12 topics and several additional labels to incoming emails, allowing customer service to filter and access them through a custom-made application. While applied to Serbian, the methodology is conceptually language-agnostic and can be readily adapted to other languages, particularly those that are low-resourced and morphologically rich. The system performance was evaluated by assessing the speed and correctness of the automatically assigned topics, with a weighted average processing time of 0.041 seconds per email and a weighted average F1 score of 0.96. The system now operates in the company's production environment, streamlining customer service operations through automated email classification.

## 1 Introduction

In recent years, there has been a significant increase in the adoption of artificial intelligence (AI) solutions by companies worldwide [1]. The growing availability of large datasets, improved computing power, and sophisticated algorithms have allowed the companies to use machine learning-based natural language processing (NLP) solutions for various purposes, leading to improved customer experience [2], minimized costs [3], and streamlined processes [1, 4].

While the NLP solutions efficiently handle routine inquiries regarding products and services, enabling the employees to focus on more complex issues, they also face several challenges, including semantic and syntactic ambiguity, managing large volumes of information, and dealing with domain-specific language [4]. These challenges are additionally intensified in low-resourced settings, where there is a scarcity of both datasets and machine learning-based tools for automating tasks.

In this paper, we present an NLP pipeline developed to improve customer service efficiency at Telekom Srbija, a major Serbian telecommunications company, by automatically detecting topics in user emails and labeling them accordingly. The central part of the pipeline is a topic detection model. The model is based on the BERTopic framework [5], which enables unsupervised clustering of semantically similar unstructured texts to form final topics. While our pipeline is currently applied to Serbian, it relies on a multilingual sentence transformer and general preprocessing and postprocessing steps, making it easily transferable to other languages. This is particularly valuable for low-resourced and morphologically rich languages, where such resources are scarce.

The rest of the paper is organized as follows. The *Related Work* section covers relevant research in the customer service domain. *Data Preparation*, *Model Training,* and *Modification of the Model Output* for purposes of implementation are presented in the *Method* section. The section *Evaluation* discusses the quality of model outputs and the applicability of the solution, while the section *Implementation* describes how the solution was applied in a real-world setting. The summary of the main findings, along with the future directions, is given in the *Conclusions and Future Work* section.

## 2 Related Work

Low-resourced languages face obstacles due to the limited availability of training data and automated processing tools [6, 7]. Most solutions in such settings rely on large multilingual models [8–12], which are adapted for specific domains.

Unsupervised methods have proven valuable in overcoming the challenges of limited training resources. While traditional approaches like LDA (Latent Dirichlet Allocation) and NMF (Non-negative Matrix Factorization) have been commonly used, in recent years, the adoption of BERTopic as a more advanced solution has been on the rise (e.g., [13, 14]). In a study related to customer service, the authors used multilingual BERT embeddings and BERTopic to analyze Turkish customer feedback and identify key complaint themes from an electronic store [10]. Their analysis revealed seasonal patterns in complaints (e.g. increased laptop complaints during distance learning), as well as monthly trends in product-specific issues (like headphone returns), demonstrating BERTopic's capability with minimal data preprocessing. Similarly, the findings from [15] have both scholarly and practical significance. The research demonstrates that BERTopic, combined with multilingual word embeddings, can effectively analyze Turkish customer reviews without requiring extensive text preprocessing, expanding its potential applications in multilingual settings.

Recently, the authors in [16] applied BERTopic to analyze customer reviews using sentence vector representations. The study examined reviews of a Russian airline service from customer websites. By comparing the automatically generated topics with the manually constructed conceptual model of the domain, the results showed strong alignment with a precision of 0.955 and recall of 0.875. This demonstrates that BERTopic effectively extracts opinion aspects from customer feedback.

Unlike previous research, the study [17] applied BERTopic to analyze customer service patents registered between 2000 and 2022, using a pre-trained language model. The researchers identified and analyzed ten key topics in customer service technology patents over time, managing to reveal the emerging topics over time. The study demonstrates how BERTopic can effectively analyze patent trends to forecast technological developments in the industry.

The mentioned findings from studies related to morphologically rich languages — Turkish [10, 15] and Russian [16], are particularly relevant to our study since they demonstrate successful topic modeling applications for feedback analysis.

## 3 Method

The development of the model for our use case involved the preparation of training data, the process of model training, and the modification of the model output.

### 3.1 Data Preparation

The dataset for model development consisted of emails provided by Telekom Srbija, collected from both individual and business users. To comply with data protection regulations and safeguard user privacy, the emails had been anonymized by replacing personal information (e.g., names, locations, organizations) with placeholders indicating their category (e.g., PER, LOC, ORG).

Since we observed that users sometimes described their problem in the email subject and left the body of the email empty or used the terms indicative of their problem only in the subject, we decided to include its content as well. After extracting the email body and concatenating it with the subject, the text of each email was preprocessed through a series of transformations designed to retain only informative and clean text, optimizing the dataset quality for model training:

1. ***Script standardization and lowercasing:*** During model training, text written in Latin and Cyrillic scripts results in separate embedding representations, even when the content is semantically identical. To address this issue, we unified the script of all emails by transliterating Cyrillic script emails into Latin. To transliterate emails, we used the Python module *strtools* [18]. All email content was lowercased to ensure that words with different capitalizations, but identical meanings have the same representation.
2. ***Punctuation, numbers, and special characters removal***: We removed punctuation, individual numbers, and special characters (e.g. symbols and emojis) since they are not informative in search of latent topics in the text.
3. ***Short document removal:*** We removed all emails containing three words or less since we found them semantically uninformative for topic modeling. The minimal number of words was determined empirically based on a manual email review.
4. ***Duplicate email removal***: Redundancy in the dataset can negatively impact the model's ability to generalize by over-representing specific patterns or topics [19].

We removed all duplicate emails to address this, ensuring that the dataset consisted only of unique emails.

5. ***Removal of closing phrases, signatures, and disclaimers***: Closing remarks (e.g., *Best regards*) appear across a wide range of emails and typically signify the end of an email and the start of signatures and disclaimers. While these elements are common, they do not contribute to identifying latent topics, as their presence is not topic-specific. To address this, we carefully identified a subset of closing phrases and removed them and the following text from the emails.
6. ***Anonymization placeholder removal:*** We remove anonymization placeholders from the emails since they appear in all emails.
7. ***Long document removal:*** The BERTopic embedding model we use imposes a limit of 128 tokens on the number of input tokens. When tokenized, longer emails can exceed this limit, resulting in only a portion of the email being processed by the model. To ensure the quality of training data, we exclude emails that surpass the token limit to avoid incomplete or misleading topic representations when training the model.

We skipped the lemmatization step because the emails contain company-specific terms, which the current lemmatizers would inadequately handle. Moreover, authors in [13] have already shown that BERTopic trained on Serbian unlemmatized text obtains informative and meaningful topics.

After preprocessing, we filtered the dataset to eliminate irrelevant or noisy emails, such as spam or automated emails (e.g., out-of-office notices), which could negatively impact topic modeling. We did so by either filtering them by language, using the Python library *langid* [20], since we found that most non-Serbian emails were spam, or by discarding the emails containing phrases indicative of automated emails. The remaining emails were further manually checked.

After filtering, we were left with approximately 90,000 emails, comprising 50,000 emails from individual users and 40,000 from business users. We randomly sampled 100 emails from both user categories to create a test dataset. The remaining emails were used for model training. The test dataset was then employed upon model training to assess the model's performance by manually comparing the assigned topics with the email content.

### 3.2 Model Training

The training was performed in an unsupervised way using BERTopic, a state-of-the-art topic modeling framework. BERTopic has a modular architecture that consists of several sequential layers that can be modified to perform the following processes: 1) extracting the embeddings, i.e. forming a dynamic vector representation of the text, 2) reducing the dimensionality of the vector representation, 3) clustering the embeddings into topics and 4) creating a representation of topics in the form of keywords. An optional step is reducing the number of outliers, i.e. documents that were not initially grouped into any of the discovered topics. We iteratively tested different architecture configurations to achieve the best possible results on our dataset. The results achieved after each change were manually analyzed against the produced topic keywords, number of topics, top representative documents, and 100 test documents.

At its first layer, BERTopic uses a sentence transformer [21], a model that significantly aids the clustering task [22], as its default embedding model. As no sentence transformer is trained only on Serbian text, we used one of the multilingual models trained on parallel data for 50+ languages, including Serbian. Following the findings of [13], we chose *paraphrase-multilingual-mpnet-base-v2*, which, in their study, produced the best results for preprocessed unlemmatized text in Serbian.

We used UMAP, the default BERTopic dimensionality reduction algorithm, to reduce the embedding dimensionality. After testing different parameter combinations, we obtained the best results using the default settings, adding the *random_state* parameter to ensure reproducibility of the model results.

BERTopic uses HDBSCAN by default to cluster the reduced embeddings. While we tested different HDBSCAN parameters and k-Means for clustering purposes, the default parameters of HDBSCAN, in combination with the BERTopic parameter *min_topic_size* set to 100, produced the best results.

By default, BERTopic uses CountVectorizer and c-TF-IDF to extract topic representations from previously created groups of documents. We tested different parameters for CountVectorizer to filter the words representing the topic (keywords). The best results, i.e. the most informative representations, were obtained by removing stop words and words that appear in less than 20 documents. At the same time, c-TF-IDF was used as provided by the BERTopic framework. As for the list of stop words, we used the expanded version of the Serbian stop word list [23], to which we further added the words that frequently appear in the training set but are not informative for topic representations. The expansion of the list was done iteratively through a manual review of the keywords obtained after training the model (e.g., *dear* or *hello*).

When used with HDBSCAN, BERTopic creates outliers, i.e., documents that are not initially grouped into any cluster that represents topics. Since there can be a significant number of documents identified as outliers, BERTopic provides the option to reduce them. We tested the *reduce_outliers* method with available strategies and finally selected the c-TF-IDF strategy, which calculates c-TF-IDF representation for each outlier document and assigns it the topic with the best matching c-TF-IDF topic representation using cosine similarity. By way of illustration, the first model output yielded around 47,000 documents as outliers, and after applying the reduction strategy, this number decreased to around 470.

As for the other BERTopic parameters, we tested *nr_topics*, which allows us to set the desired number of topics. Although it is recommended not to limit the model with this parameter, we tested its different values. By manually reviewing the keywords of topics obtained this way, we determined that certain topics were too general or "mixed", i.e. that they did not separate well the topics that the customer service needed as distinct. For example, when we set 15 as the number of topics, almost 90% of the emails were grouped into one topic.

Another parameter we tested was *min_topic_size*, which defines the minimum number of documents that must be grouped for the model to declare a cluster. We tested various values of this parameter, from default 10 to several hundred. As defining a lower value led to the creation of a large number of topics (in some cases over 400), and defining a high value led to fewer topics containing topics that needed to be separate (as was the case with *nr_topics*), we set this number to 100 for our purposes.

BERTopic provides guided topic modeling using the optional *seed_topic_list* parameter. This option allows defining seed topics through a list of predefined topic representations, i.e. keywords, to which the model will attempt to converge. This parameter was tested by creating several different lists of keywords to refine certain topics. However, no significant improvement was achieved except for one topic, for which we decided to keep the seed topic list.

The model was trained in an environment with Python 3.10.10, using version 0.14.1 of the BERTopic library [24], which was current at the time.

### 3.3 Modification of the Model Output

The most optimal BERTopic configuration described in the subsection *Model Training* led to 72 topics and an additional outlier group. As 72 topics are a vast number for customer service applications, we conducted a manual analysis of the topic keywords and representative emails for each topic to determine the optimal number of topics aligned with the company's services and common user concerns. This way, we detected that all 72 topics could fall into 12 clearly defined groups, which the customer service team approved as meaningful. This meant that we needed to reduce the current optimal number of topics from 72 to this number. As testing the model parameters *min_topic_size* and *nr_topics* to automatically generate 12 topics did not yield the same ones we detected, we needed to resort to a different strategy.

Another strategy we tested was hierarchical topic modeling, which attempts to capture the possible hierarchical structure of the generated topics from the c-TF-IDF matrix to identify which topics are similar and could be merged. To analyze both the general structure and the hierarchy of topic representations in more detail, as well as to evaluate the effects of merging certain topics, we used the *visualize_hierarchy()* and *get_topic_tree()* methods. Following manual analysis, we applied the *merge_topics()* method to reduce the number of topics. This process updates the topic representations of the merged topics and, subsequently, the entire model. This method proved effective for the initial couple of merges. However, as the number of merges rose, the method started to produce unsatisfactory results.

Finally, we resorted to the manual grouping of the model topics. BERTopic allows the users to add custom labels for each topic. We used this option as a topic reduction strategy by assigning the same 12 custom labels to all the original topics we grouped together, creating the *derived topics*. In addition, we analyzed the outlier group and assigned one of the derived topics to it for practical purposes, naming it *General problems and malfunctions*. Since this strategy does not influence the initial clusters of the model, it allows for further modifications of the derived topics if required by customer service. An example of the mapping of original topics into a derived one can be seen in **Table 1**.

**Table 1.** The mapping of 13 original topics, represented by their 10 representative keywords, to a derived topic.

| Derived topic | Original topic | Keywords |
|---|---|---|
| **Računi i fakture (*Bills and invoices*)** | 30 | ['dinara', 'iznos', 'objasnite', 'din', 'cena', 'račun', 'iznosu', 'racun', 'naplaćuje', 'iznosi']<br>*['dinar', 'amount', 'explain', 'din', 'price', 'bill', 'amount', 'bill', 'charges', 'amounts']* |
| | 70 | ['dobili', 'stigao', 'poštom', 'račun', 'pošaljete', 'mesec', 'račune', 'racun', 'mail', 'dostavite']<br>*['received', 'arrived', 'by post', 'bill', 'send', 'month', 'bills', 'bill', 'e-mail', 'deliver'*] |
| | 4 | ['dugovanje', 'dug', 'dugovanja', 'izmirili', 'duga', 'odnosi', 'izmirena', 'račun', 'iznosu', 'perioda']<br>*['owing', 'debt', 'owing', 'settled', 'debt', 'refers to', 'settled', 'bill', 'amount', 'period']* |
| | 60 | ['dva', 'platila', 'greskom', 'racun', 'platio', 'racuna', 'isti', 'uplatila', 'račun', 'uplatu']<br>*['two', 'paid', 'by mistake', 'bill', 'paid', 'bill', 'the same', 'paid', 'bill', 'payment']* |
| | 43 | ['faktura', 'fakture', 'plaćanje', 'fakturu', 'br', 'iznos', 'sistemu', 'crf', 'sistem', 'računa']<br>*['invoice', 'invoices', 'payment', 'invoice', 'no', 'amount', 'system', 'crf', 'system', 'invoice']* |
| | 18 | ['mesec', 'pošaljete', 'dostavite', 'posaljete', 'kopiju', 'račun', 'račune', 'racune', 'računa', 'firmu']<br>*['month', 'send', 'deliver', 'send', 'copy', 'bill', 'bills', 'bills', 'bills', 'company']* |
| | 71 | ['naknada', 'pretplata', 'din', 'mesečna', 'mesecna', 'cene', 'iznosi', 'cena', 'naknade', 'rsd']<br>*['fee', 'subscription', 'din', 'monthly', 'monthly', 'prices', 'amounts', 'price', 'fees', 'rsd']* |
| | 67 | ['pdv', 'prethodnog', 'potpisano', 'om', 'predhodnog', 'din', 'potpisana', 'dinara', 'račun', 'storno']<br>*['VAT', 'previous', 'signed', 'om', 'previous', 'din', 'signed', 'dinar', 'invoice', 'canceled']* |
| | 57 | ['placen', 'racun', 'racuna', 'mesec', 'racuni', 'uplaceno', 'uplacen', 'uplata', 'placeni', 'uplatnicu']<br>*['paid', 'bill', 'bill', 'month', 'bills', 'paid', 'paid', 'payment', 'paid', 'payment slip']* |
| | 15 | ['poziv', 'uplata', 'račun', 'uplatu', 'uplate', 'plaćen', 'racuna', 'računa', 'broj', 'racun']<br>*['call', 'payment', 'account', 'payment', 'payments', 'paid', 'account', 'account', 'number', 'account']* |
| | 68 | ['račun', 'iznos', 'računa', 'iznosu', 'mesec', 'rsd', 'greškom', 'računu', 'prigovor', 'dinara']<br>*['bill', 'amount', 'bill', 'amount', 'month', 'rsd', 'by mistake', 'bill', 'objection', 'dinar']* |

| | |
|---|---|
| 41 | ['račune', 'šaljete', 'adresu', 'mail', 'ubuduće', 'mejl', 'računi', 'račun', 'računa', 'formi']<br>*['bills', 'send', 'address', 'e-mail', 'in the future', 'e-mail', 'bills', 'bill', 'bills', 'form']* |
| 10 | ['uplati', 'dokaz', 'izvod', 'ukljucenje', 'uplata', 'uplatu', 'računa', 'racuna', 'uplate', 'uključenje']<br>*['payment', 'proof', 'statement', 'reconnection', 'payment', 'payment', 'bill', 'bill', 'payments', 'reconnection']* |

## 4 Evaluation

The model was primarily evaluated for the functionalities needed in the implementation environment, namely processing speed and assignment correctness.

The first parameter was the time needed for a topic assignment. The processing time span for the entire topic assignment pipeline for a batch of 100, 1,000, and 10,000 emails across three runs was 0.035-0.112 seconds per email, with a weighted average of 0.041 seconds per email. The processing time for 100 emails was consistently the highest (0.086-0.112), suggesting a slight negative effect of smaller processing batches. An increase to a batch of 1,000 emails showed significant efficiency improvements (0.035-0.038), which once again started to decline with a batch of 10,000 emails (0.040-0.041). Despite these variations, the increase in processing time between 1,000 and 10,000 emails is roughly linear, showing that the model scales well with more data.

The correctness of the topic assignment was evaluated by manually reviewing the assigned topics for the selected 100 test emails. The emails were first reconstructed to include fictitious personal names, locations, and organization names instead of the previously anonymized information to mimic real-life correspondence. They were then assigned one of the 12 derived topics based on the primary concern in the email, after which this topic was compared to the one assigned by the model. In 44 cases, the email contained more than one topic. We identified two types of these situations, the one in which two or more topics have equal significance and the other in which there is one more dominant topic. For emails containing topics with equal relevance (7 cases), we considered the topic correct if it matched any of the topics we identified in the email. For the remaining 37 cases with one dominant topic, we considered the assigned topic correct only if it matched the topic we selected. The results of this evaluation, expressed by accuracy, weighted average precision, recall, and F1 score, can be seen in **Table 2**.

**Table 2.** Topic model evaluation results on 100 test emails.

| Metric | Result |
|---|---|
| Accuracy | 0.9600 |
| Weighted average precision | 0.9645 |
| Weighted average recall | 0.9600 |
| Weighted average F1 | 0.9599 |

By more detailed manual error analysis, we observed that the topic was incorrectly assigned in four emails, where the model was usually misled by the appearance of words that are key representatives of other topics. In one of those cases, the assigned topic is correct but not the dominant one. This means that the model correctly assigned the dominant topic to most emails with more than one topic. In situations with topics of equal significance, the model always correctly assigned one of those topics.

To account for the fact that emails can contain more than one topic, we also evaluated different ways for the model to detect an additional topic in an email, even though, by design, it assigns only one topic to one document. One of the ways was to access the probabilities of each document belonging to all topics using the *calculate_probabilities* parameter of the model and then extracting the second highest probability. If the second highest probability belongs to a topic grouped under the same derived topic, we would assume there is only one main topic in the document. Otherwise, the topic with the second highest probability would be assigned as the second one. However, the test results showed that this strategy is not accurate enough for most of the observed cases. We also tested the solution that approximates topic distributions using the *approximate_distribution()* method. Although it is not as accurate for prediction as the *transform()* method that is usually used [25], we wanted to test if the topic with the second highest probability obtained this way could be the additional topic. We gained better results compared to the first solution, but they were still not satisfactory enough to be a reliable choice for practical application.

We additionally evaluated the topic assignment for longer emails. The embedding layer of the model allows topic assignment for a maximum of 128 tokens per document. Therefore, any email longer than this is reduced by truncating the rest of the text. We tested the possibility of splitting emails longer than 128 tokens, assigning a topic to each segment, and then deciding on the final topic, based either on the most frequently assigned derived topic or the topic with the highest probability. This method did not give satisfactory results, leaving the challenge of handling longer emails for future work. However, our analysis shows that this limitation does not significantly impact the overall topic assignment performance. For reference, the statistics of emails longer than 128 tokens before long document removal (see the subsection *Data Preparation*) was 7.8% for emails from individual users and 3.5% from business users, which is a relatively low prevalence Furthermore, by manually analyzing those longer emails, we determined that 128 tokens are usually long enough for the main user message to be processed, indicating that the current approach is sufficient for most cases.

## 5 Implementation

The described topic modeling system is integrated into Telekom Srbija's production environment to enhance customer support by automating email categorization. It comprises four main components: a database, a topic model, and email preprocessing and postprocessing modules. When new emails arrive, they are stored in the database and periodically retrieved in batches for topic modeling. While the model can assign topics to emails quickly enough to support real-time topic modeling, we implemented a periodic batch processing approach to align with the company's operational requirements. The simplified workflow is shown in **Fig. 1**.

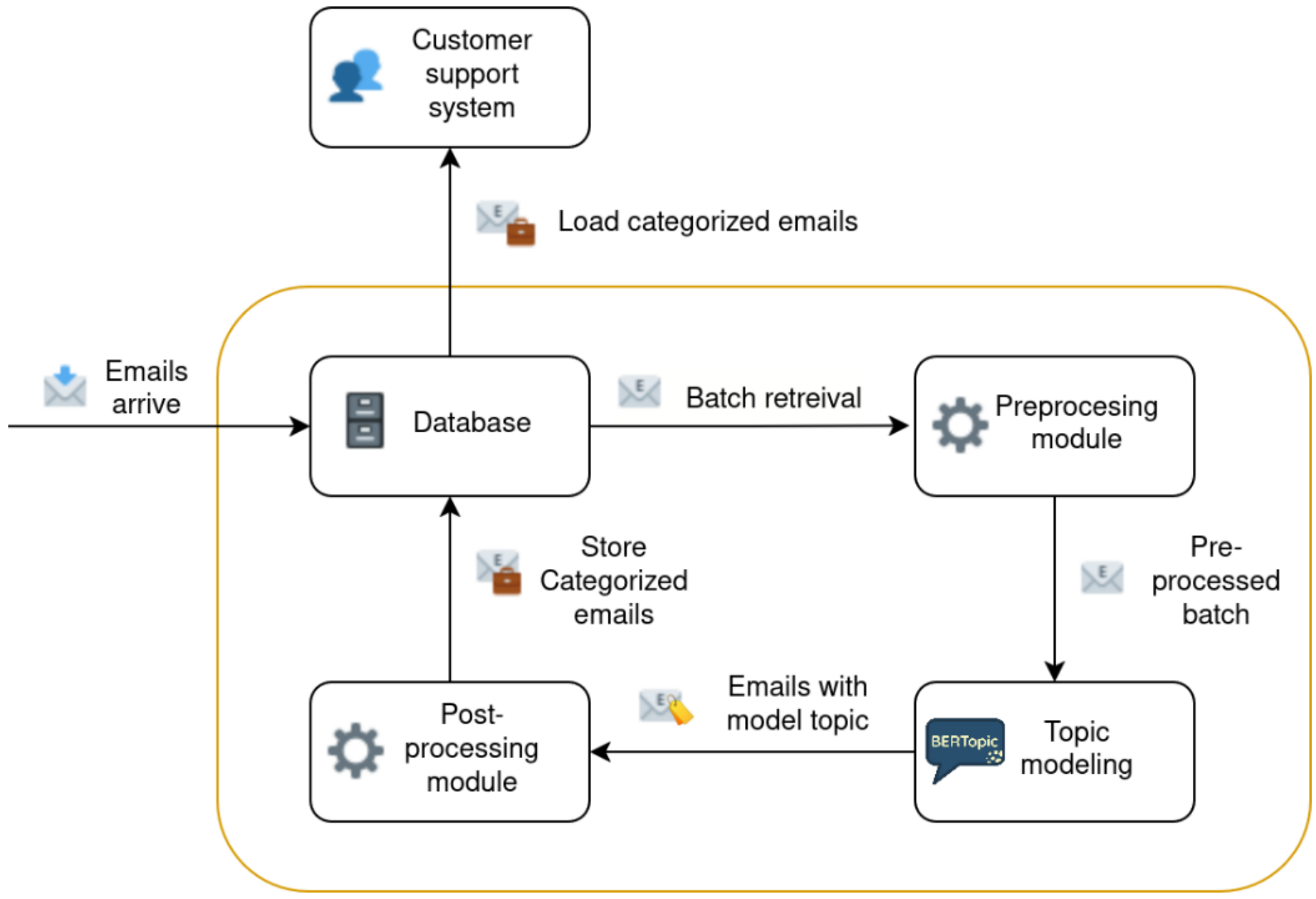

**Fig. 1.** The outline of the topic modeling workflow.

Upon retrieving a batch of emails from the database, each email is run through the preprocessing module to filter out irrelevant emails and standardize the remaining ones. The preprocessing module first filters the incoming emails based on a list of predefined email addresses to exclude internal correspondence. Such emails are automatically labeled *Internal Correspondence* and are not further processed. The remaining emails, concatenated with their subjects, go through transformations 1, 2, and 5 of the preprocessing pipeline described in the *Data Preparation* subsection, with an additional step of detecting and deleting parts of emails that are a reply or forwarded. If the outcome of preprocessing is an empty string, the email is automatically labeled *Spam, a reply, forwarded, or empty*. These emails are also excluded from the assignment process. Additionally, emails are checked for language. If the detected language is English, the email is again labeled *Spam, a reply, forwarded, or empty* and is excluded from further processing.

The topic model then assigns one of the 72 model topics to each remaining email, followed by the outlier reduction process. To integrate our model into the company's production environment, we have developed the software infrastructure that connects the model with the database that stores all incoming emails.

The postprocessing module processes emails with the assigned model topics and maps them to the 12 predefined derived topics. Finally, the derived topic is stored in the database for each processed email.

Customer service representatives use a custom-built application to access categorized emails. Through this application, they can filter emails by category of interest and quickly identify groups of concern, such as clients who are dissatisfied with the service and at risk of discontinuing company services. The structured categorization enables targeted intervention, allowing support teams to address critical issues proactively.

## 6 Conclusions and Future Work

This paper presents a topic modeling pipeline for the automatic assignment of topics to user emails. The pipeline is specifically developed to enhance customer service at Telekom Srbija, a major Serbian telecommunications company, allowing it to prioritize certain groups of emails based on the assigned topics. The pipeline relies on BERTopic, which assigns the customer-service-approved topic to an email after a series of preprocessing and postprocessing steps. Since the underlying embedding model is multilingual and the pipeline requires minimal language-specific pre- and postprocessing, it can be easily adapted to other languages. Based on the weighted average processing time of 0.041 seconds per email and the correctness of topic assignment at a weighted average F1 score of 0.96, the system proves effective and reliable for the given purpose. The topics and labels for the filtered emails (e.g., internal correspondence and spam) are stored in a database, which further allows the company to create monthly reports and analyze the prevalence of certain customer concerns over time.

For future work, we plan to enhance the pipeline to accurately assign multiple topics per email and process messages beyond the current 128-token limit. Assigning multiple topics is important, as customer emails may address multiple issues that require attention from multiple customer service departments. In larger companies, these departments are often separate, making it necessary to route emails correctly to ensure timely responses. Similarly, while we determined that the size of 128 tokens is enough to capture the main concern in an email, we still plan to research ways to process entire longer emails. The planned improvements would enhance system accuracy and ensure better coverage of incoming emails.

Furthermore, as company services evolve, new customer concerns may emerge. Integrating a mechanism to flag potential novel topics would address this challenge and further allow for either grouping of such emails into a separate topic or call for model retraining. Another possible future work challenge is handling ambiguous emails. While humans can ask for clarifications when faced with ambiguous messages, automated systems need to resort to alternative strategies. Introducing a confidence score for topic predictions and flagging low-confidence predictions can serve as a sign for customer service to pay closer attention to such emails.